\documentclass{vgtc}                          % final (conference style)
\graphicspath{{figures/}{pictures/}{images/}{./}} % where to search for the images

\usepackage{times}                     % we use Times as the main font
\usepackage{tabu}                      % only used for the table example
\usepackage{booktabs}                  % only used for the table example
\usepackage{lipsum}                    % used to generate placeholder text
\usepackage{mwe}                       % used to generate placeholder figures

\usepackage{mathptmx}                  % use matching math font

\usepackage{subcaption}
\usepackage{dblfloatfix}

\onlineid{0}

\vgtccategory{Research}

\vgtcinsertpkg

\title{Cross-Platform Benchmark of Neural 3D Reconstruction for Autonomous Laboratory Robots}

\author{Yongho Kim\thanks{e-mail: yongho.kim@anl.gov}, Mengjiao Han\thanks{e-mail: hanm@anl.gov} \\ %
    \scriptsize Argonne National Laboratory %
\and Victor Mateevitsi\thanks{e-mail: vmateevitsi@anl.gov}, Silvio Rizzi\thanks{e-mail: srizzi@anl.gov}, Michael E. Papka\thanks{e-mail: papka@anl.gov, papka@uic.edu}\\ %
    \scriptsize Argonne National Laboratory \\ \scriptsize University of Illinois Chicago %
\and Nicola Ferrier\thanks{e-mail: nferrier@anl.gov}\\ %
    \scriptsize Argonne National Laboratory \\ \scriptsize NAISE, Northwestern University %
}

\teaser{
  \centering
  \includegraphics[width=\linewidth]{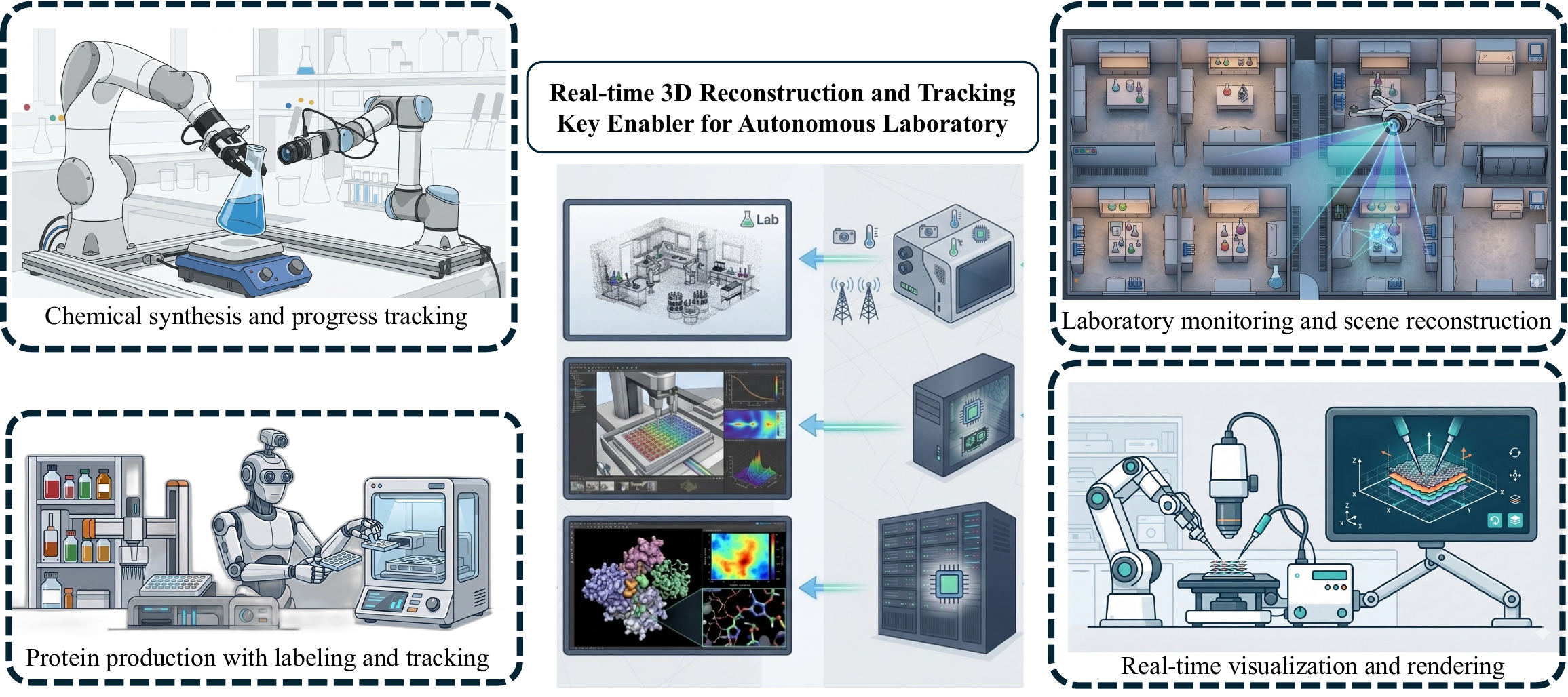}
  \caption{Our vision is an autonomous laboratory powered by 3D reconstruction and tracking. The four dashed boxes illustrate use cases necessitating visual analytics for science-aware autonomous manipulation. Note that the illustration is AI-generated.}
  \label{fig:teaser}
}

\abstract{
    Autonomous robots performing laboratory tasks depend on 3D reconstruction pipelines that can turn raw camera streams into actionable object representations within the latency budget of a physical control loop. Neural 3D reconstruction methods have demonstrated high-quality view synthesis, but their real-time viability across the compute platforms on which laboratory robots actually run remains poorly characterized. In this work, we present a systematic compute-platform benchmark of neural 3D reconstruction methods, evaluating NeRF and 3D Gaussian Splatting training and rendering on GPU-enabled computing devices ranging from single-board computers to server-class nodes, and place Meta's SAM3D single-image reconstruction on the same axes to quantify its latency and fidelity gap relative to per-scene optimization. Our results show that Gaussian Splatting yields higher rendering quality than NeRF at greater GPU cost, and that onboard compute is insufficient for full per-scene optimization at interactive rates. Our preliminary assessment on SAM3D indicates that it delivers plausible object geometry within seconds, but with detail mismatches that can compromise downstream manipulation. Together, these findings motivate tiered pipelines in which lightweight feed-forward reconstruction sustains the real-time perception-and-tracking loop for laboratory robots, while heavier neural reconstruction is scheduled selectively on suitable compute.
} % end of abstract


\begin{document}

%% The ``\maketitle'' command must be the first command after the
%% ``\begin{document}'' command. It prepares and prints the title block.
%% the only exception to this rule is the \firstsection command
% \firstsection{Introduction}
\maketitle
\section*{Acknowledgement}
This work was supported in part by the Office of Science of the U.S. Department of Energy under Contract No. DE- DE-AC01-06CH11357 through two projects: 1) "DAIMSL: \textbf{D}igital twins and \textbf{A}I-enabled  \& \textbf{IM}mmersive \textbf{E}nvironments for Automated \textbf{S}cientific \textbf{L}aboratories''. funded by the Office of Advanced Scientific Computing Research (ASCR),  and 2)``Tele-robotics to Tele-autonomous robotics for isotope production'', a project jointly funded by the Office of Isotope R\&D and Production (IP) and the Office of Advanced Scientific Computing Research (ASCR). This research used resources of the Argonne Leadership Computing Facility, which is a DOE Office of Science User Facility supported under Contract DE-AC02-06CH11357. The manuscript is written and edited with AI assistance.

% \section*{Government License}
The submitted manuscript has been created by UChicago Argonne, LLC, Operator of Argonne National Laboratory (``Argonne"). Argonne, a U.S. Department of Energy Office of Science laboratory, is operated under Contract No. DE-AC02-06CH11357. The U.S. Government retains for itself, and others acting on its behalf, a paid-up nonexclusive, irrevocable worldwide license in said article to reproduce, prepare derivative works, distribute copies to the public, and perform publicly and display publicly, by or on behalf of the Government. The Department of Energy will provide public access to these results of federally sponsored research in accordance with the DOE Public Access Plan (http://energy.gov/downloads/doe-public-accessplan).

\section{Introduction}
Autonomous laboratories are increasingly relying on mobile manipulators and instrument-mounted robots to carry out experimental tasks that once required a human scientist at the bench --- loading samples, operating centrifuges, transferring liquids, and inspecting apparatus~\cite{ming2025benchmarking}. Each of these tasks depends on a 3D representation of the target object that is accurate enough for reliable manipulation and fast enough to keep the robot's perception–action loop. 3D reconstruction and object tracking are therefore not peripheral capabilities in a scientific robot's software stack; they sit directly on the critical path between sensing and action.

Neural 3D reconstruction has emerged as a natural candidate for this role. Neural Radiance Fields (NeRFs) and 3D Gaussian Splatting (GS) produce dense, photorealistic, view-consistent reconstructions from multi-view images, while transformer-based large reconstruction models (LRMs) such as SAM3D generate a 3D object from a single image in seconds. These two families sit at opposite ends of a latency–fidelity spectrum: per-scene optimization delivers high-quality geometry at the cost of minutes to hours of training, while feed-forward inference delivers near-instantaneous outputs whose fidelity, especially in occluded regions, is not guaranteed to match the physical object a robot must manipulate. Which family—and which compute platform—is appropriate for a given laboratory task is not obvious, and empirical guidance grounded in the hardware that scientific robots actually run on is limited.

In this work, we systematically benchmark existing neural 3D reconstruction methods on the compute tiers relevant to autonomous laboratory deployment. We evaluate NeRF and GS training and rendering, using the nerfacto and splatfacto implementations in NerfStudio, on an embedded NVIDIA Jetson AGX Orin, an RTX-class desktop workstation, and an A100-equipped HPC node, using a 40-image capture of a robotic arm as the target scene. In parallel, we present a preliminary assessment of SAM3D single-image reconstruction on laboratory objects, including a centrifuge bucket, and discuss the methodological challenges of comparing single-image LRM outputs against multi-view ground truth. Together, these results characterize the trade-offs between reconstruction quality, training and rendering latency, and compute cost that are critical to shape the design of 3D reconstruction pipelines for autonomous laboratory robots; visual analytics capabilities for science-aware decision-making are identified as future work.

\begin{figure}[tb]
\centering
  \begin{subfigure}[b]{0.26\columnwidth}
    \centering
    \includegraphics[width=\columnwidth, height=\columnwidth]{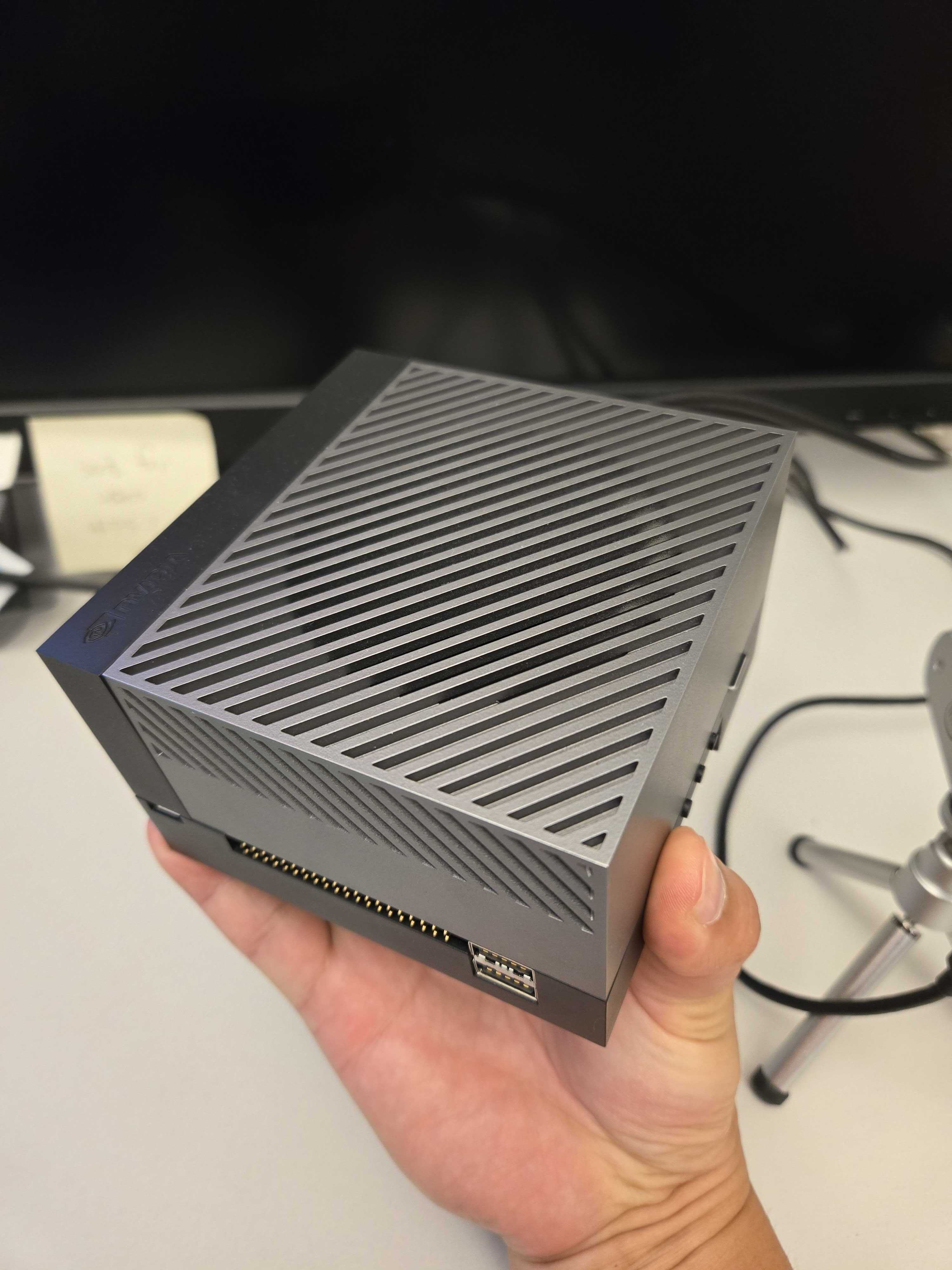}
    \label{fig:device_orin}
  \end{subfigure}
  \hfill
  \begin{subfigure}[b]{0.26\columnwidth}
    \centering
    \includegraphics[width=\columnwidth, height=\columnwidth]{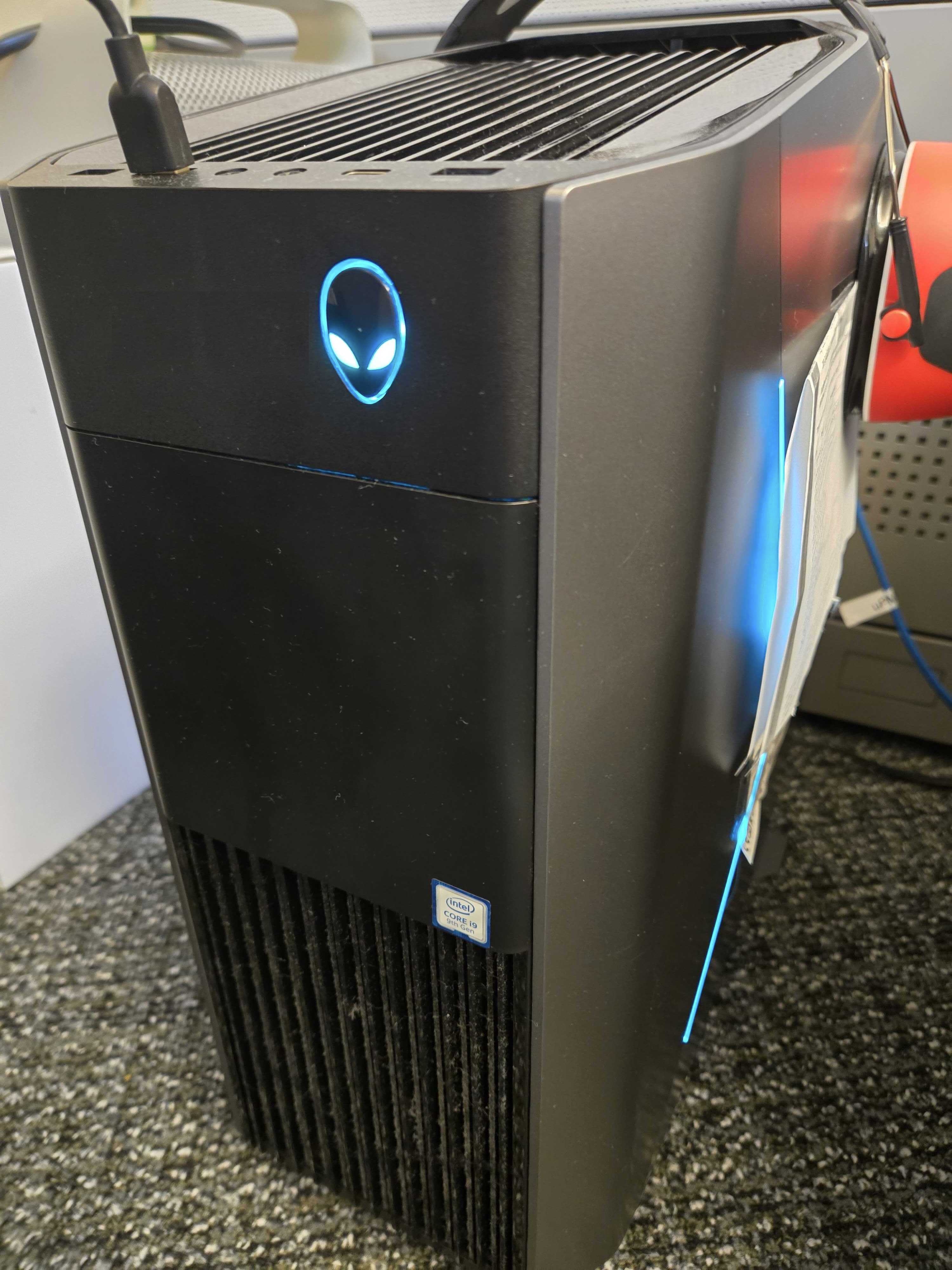}
    \label{fig:device_alienware}
  \end{subfigure}
  \hfill
  \begin{subfigure}[b]{0.46\columnwidth}
    \centering
    \includegraphics[width=\columnwidth]{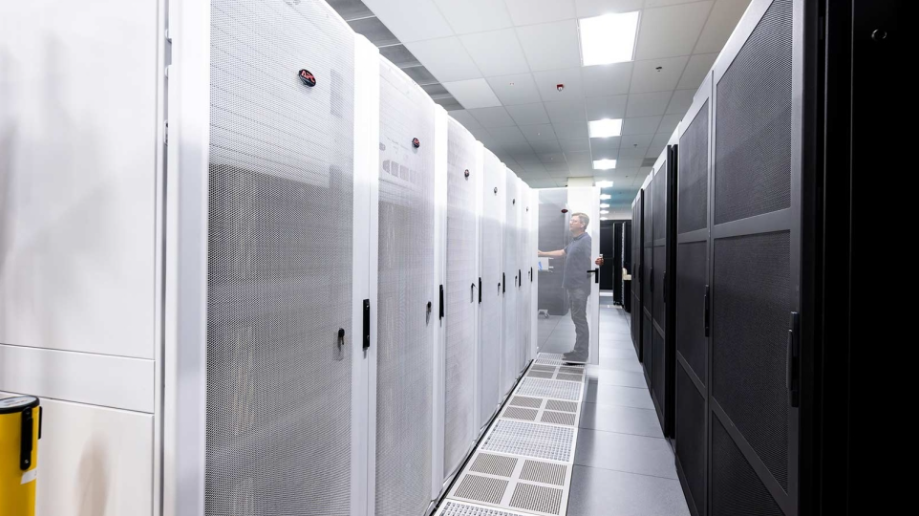}
    \label{fig:device_sophia}
  \end{subfigure}
  \caption{Computing devices used in the benchmark. (left) NVIDIA Jetson AGX Orin, (middle) Dell Alienware with NVIDIA RTX 2080, and (right) Argonne Leadership Computing Facility's Sophia with NVIDIA A100.}
  \label{fig:devices}
\end{figure}

\begin{figure}[tb]
\centering
  \includegraphics[width=\columnwidth]{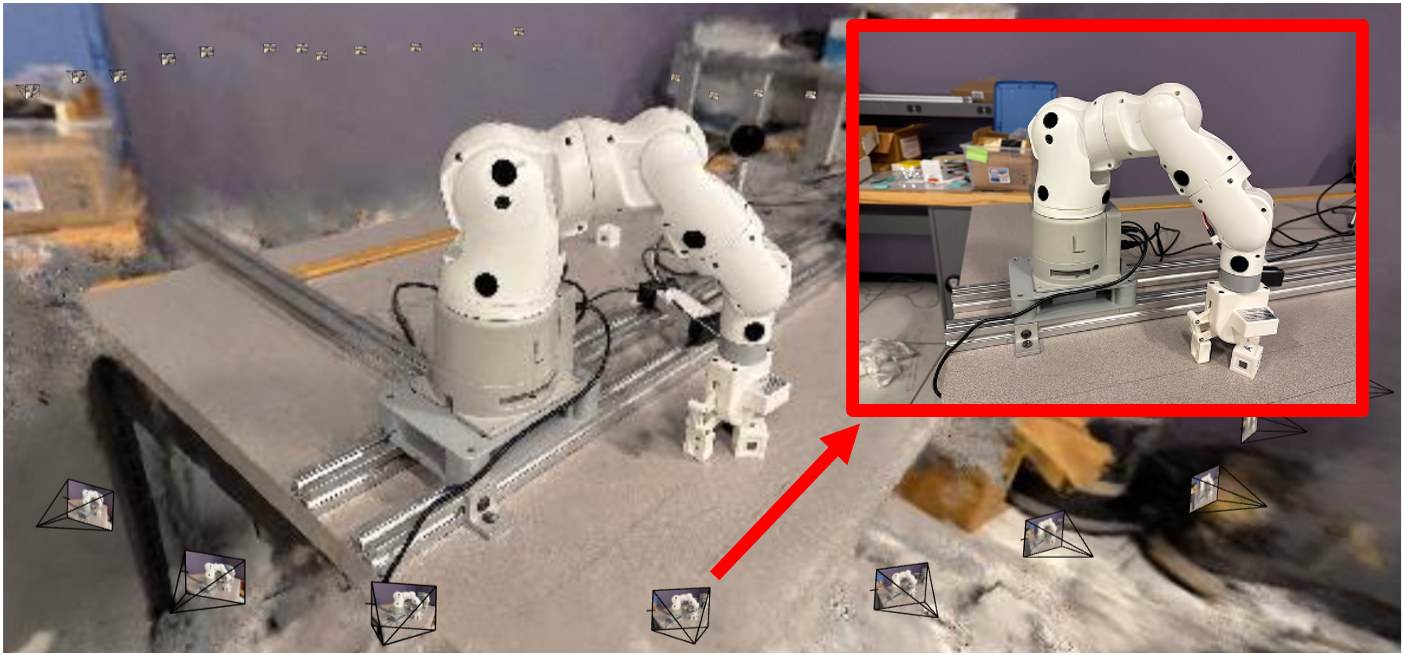}
  \caption{A synthetic view representing the scene we use for benchmarking. A total of 40 images with a resolution of 2124 x 2832 are used to train NeRF and GS models.}
  \label{fig:robotdataset}
\end{figure}

\section{Related Work}
Real-time novel-view synthesis and object tracking together constitute the core problem discussed in this paper, and both have advanced rapidly along two largely independent research fronts. On the reconstruction side, Neural Radiance Field (NeRF)~\cite{mildenhall2021nerf} and 3D Gaussian Splatting (GS) have become the dominant tools for high-fidelity static scene reconstruction, offering photorealistic novel-view synthesis at the cost of per-scene optimization. In parallel, transformer-based large reconstruction models (LRMs) have emerged, employing feed-forward neural networks to produce 3D object representations directly from one or a few images, trading fidelity for near-instantaneous inference. These two families occupy opposite ends of the latency–fidelity spectrum, and each imposes markedly different demands on the compute platform that hosts it—an axis that becomes decisive once the target deployment is a scientific robot rather than an offline workstation. The remainder of this section reviews each family in turn: we first survey their rendering and tracking capabilities, and then examine the computational requirements that govern whether, and where, they can be deployed in a robot-driven discovery loop.

\begin{figure*}[tb]
\centering
  \begin{subfigure}[b]{0.33\textwidth}
    \centering
    \includegraphics[width=\textwidth]{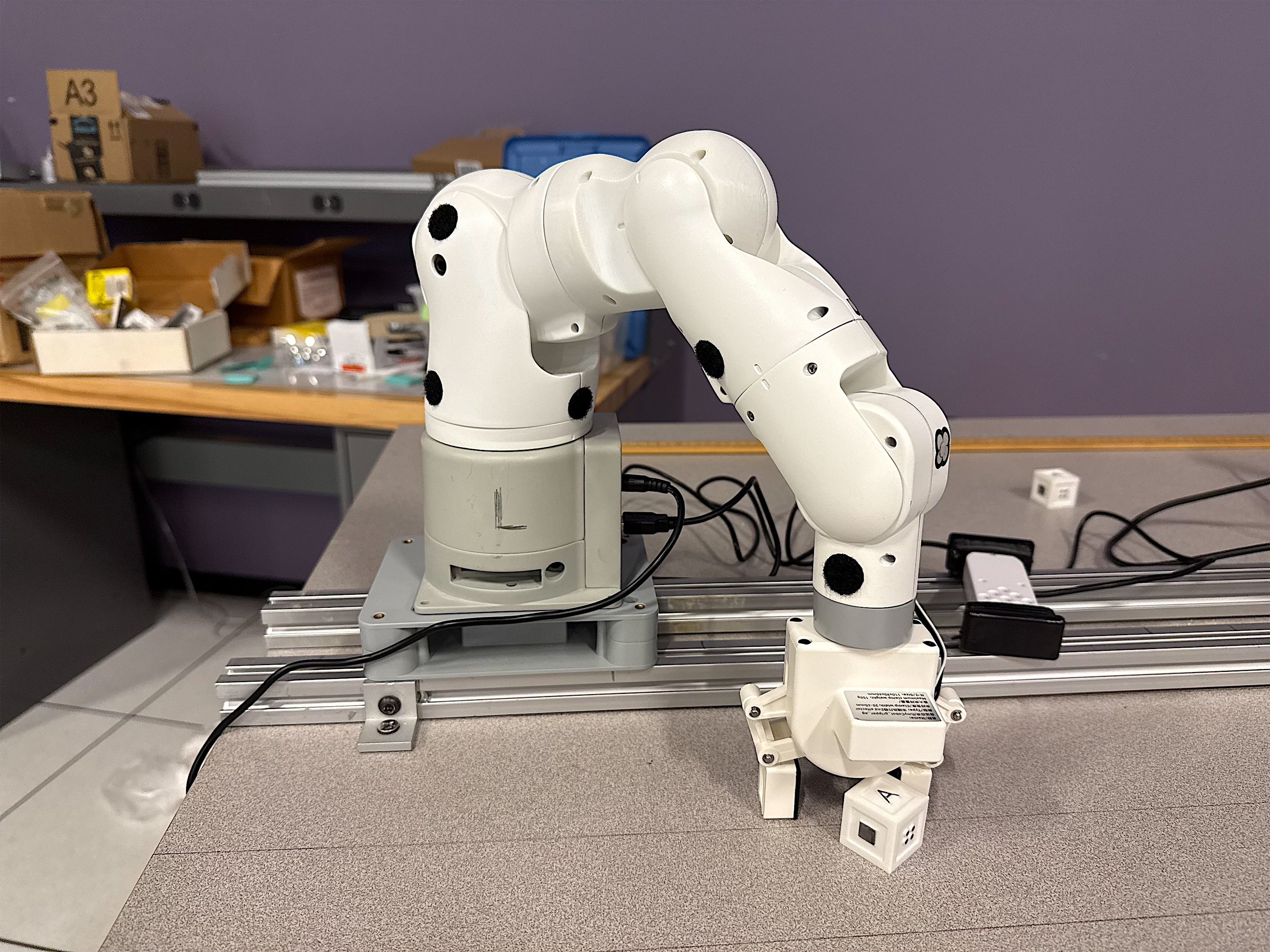}
    \label{fig:device_orin}
  \end{subfigure}
  \hfill
  \begin{subfigure}[b]{0.33\textwidth}
    \centering
    \includegraphics[width=\textwidth]{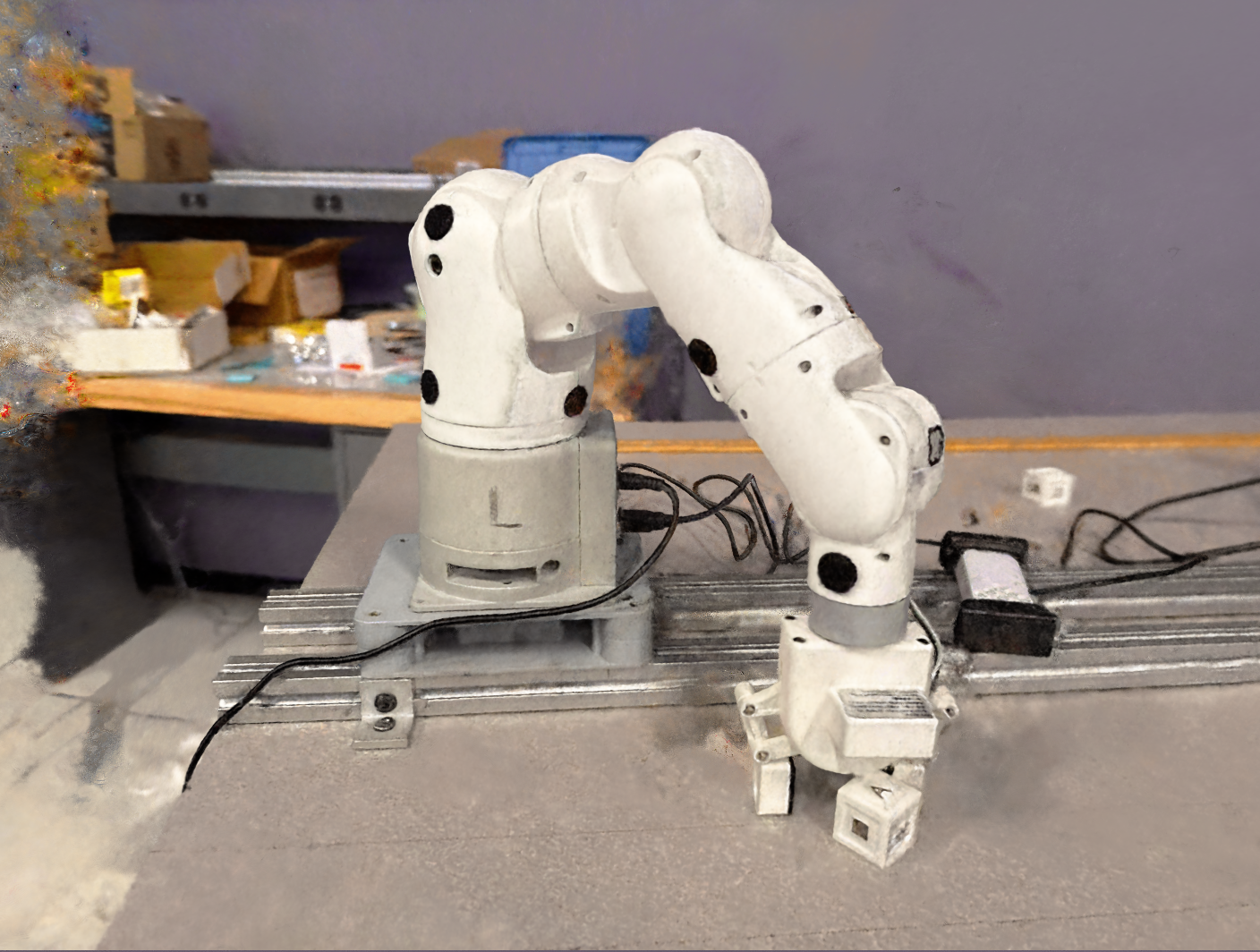}
    \label{fig:device_alienware}
  \end{subfigure}
  \hfill
  \begin{subfigure}[b]{0.33\textwidth}
    \centering
    \includegraphics[width=\textwidth]{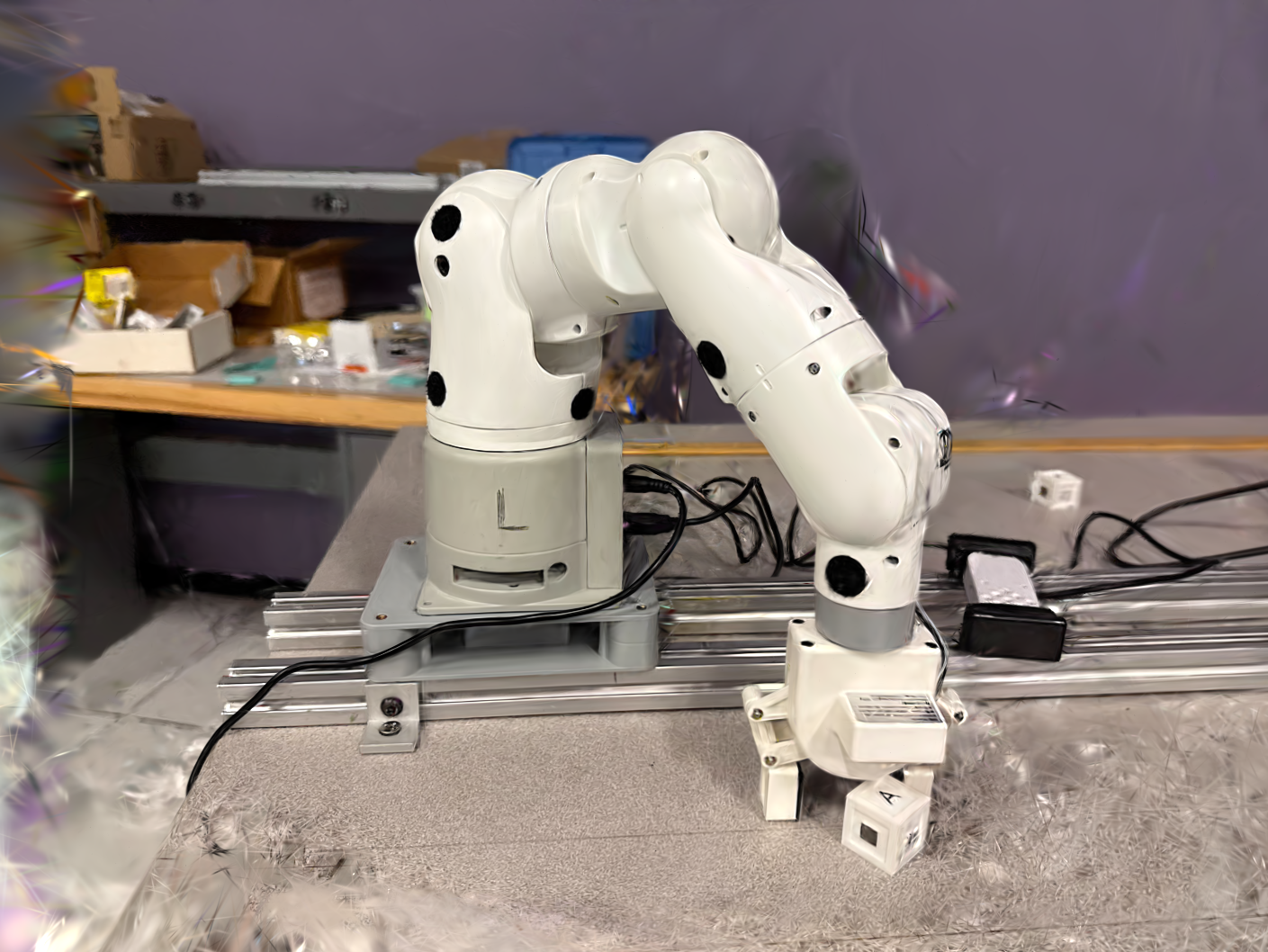}
    \label{fig:device_sophia}
  \end{subfigure}
  \caption{Visualization of the evaluation. (left) ground truth, (middle) NeRF, and (right) GS.}
  \label{fig:roboteval}
\end{figure*}

\begin{table*}[tb]
\centering
\caption{Performance Metrics of NeRF and GS across devices.}
\label{tab:benchmark}
\begin{tabular}{lccccccc}
\hline
\textbf{Case} & \textbf{PSNR $\uparrow$} & \textbf{SSIM $\uparrow$} & \textbf{LPIPS $\downarrow$} & \textbf{FPS $\uparrow$} & \textbf{Time} (minute) & \textbf{Averaged GPU Utilization} & \textbf{GPU-minute}\\ \hline
ALCF Sophia-NeRF        & 18.96 & 0.65 & 0.36  & 0.25  & 15.6  & 51.9 & 8.1 \\
ALCF Sophia-GS          & 24.32 & 0.84 & 0.21  & 0.78  & 16.5  & 83.9 & 13.8 \\
Alienware-NeRF          & 20.37 & 0.69 & 0.30  & 0.42  & 17.4  & 62.7 & 10.9 \\
Alienware-GS            & 23.21 & 0.82 & 0.21  & 24.16 & 21.8 & 72.2 & 15.7 \\
NVIDIA Jetson Orin-NeRF & 19.14 & 0.64 & 0.33  & 0.09  & 84.1  & 62.2 & 52.3 \\
NVIDIA Jetson Orin-GS   & 24.06 & 0.83 & 0.21  & 4.87  & 90.0  & 72.6 & 65.3 \\ \hline
\end{tabular}
\end{table*}
\subsection{Neural Radiance Field and Gaussian Splatting}
While NeRF reconstructs the scene using ray-tracing that passes through the network weights, GS maintains explicit ``splats'' to project onto the 2D image plane. Because this projection can be highly accelerated by a GPU, the GS method has been mainly used for rendering scenes. Instant-NGP~\cite{muller2022instant} is one of the early GS methods that utilizes multi-resolution hash encoding and pushes the training and rendering into seconds to minutes from hours. Followed by that, dynamic 3DGS~\cite{luiten2024dynamic} and 4DGS~\cite{wu20244d} extend the reconstruction to dynamic scenes for movement tracking while benefiting from real-time rendering and high-quality view synthesis. To support live camera streams from mobile robots like drones flying over the scene, NerfBridge~\cite{yu2023nerfbridge} provides a messaging layer that flows image data into the training and constantly updates the image pool used in the reconstruction with unseen images for diversifying the views. 3DGauCIM~\cite{huang20263dgaucim} proposes memory-efficient algorithms for 3DGS to achieve real-time rendering in power-constrained edge devices.

However, we practically experience that the training of these GS models can take at least a few minutes to over an hour to obtain a high-quality reconstruction. For instance, 4DGS~\cite{wu20244d} uses NVIDIA RTX 3090 and takes 8 minutes of training, still beating many existing GS methods. Additionally, the training requires mutli-perspective 2D images. These can become a constraint when needing a within-a-minute reconstruction and tracking for robots to manipulate science objects in real time.

\subsection{Large Reconstruction Models}
Large reconstruction models are transformer-based models for reconstructing objects using a single image within seconds~\cite{hong2024lrm}. Once trained with millions of 3D objects, the model predicts NeRF tri-plane representations with volumetric rendering to construct a 3D model. LRMs are widely used to construct object meshes and 3D Gaussian primitives for reconstruction~\cite{wei2024meshlrm, zhang2024gs}. SAM3D~\cite{chen2026sam}, possibly fused with SAM3 object tracking~\cite{carion2025sam}, and BundleSDF~\cite{wen2023bundlesdf} maintain their memory pool to save posed models to track objects in a video. The large 4D Gaussian reconstruction model~\cite{ren2024l4gm} generates an animated 3D object from a video. Note that they are not necessarily all transformer models but have self-attention layers in common.

As one can guess, one obvious downside of LRMs is the fact that the generated 3D models contain parts that are not from the input image but are predicted based on the trained model. These unseen, synthetically generated parts may fail to represent the actual object shape, possibly leading to failure in object manipulation due to the mismatch between the generated object model and the physical object. To overcome this limitation, robots can take additional perspective images to reinforce the generated model before manipulating the object (i.e., BundleSDF).

\section{Neural reconstruction on heterogeneous compute platforms}
While the aforementioned neural 3D reconstruction models push on digitizing physical objects and scenes, we focus in this work on evaluating these models in different computing devices that vary in computing architecture and energy budget. As shown in Figure ~\ref{fig:devices}, the considered devices include NVIDIA Jetson AGX Orin, a desktop-grade computer with NVIDIA RTX, and a single HPC node equipped with NVIDIA A100 GPU. Based on their specification, they consume roughly 50 W, 500 W, and 1500 W, respectively.

\subsection{NeRF and Gaussian Splatting}
We use NerfStuido~\cite{tancik2023nerfstudio}, a framework for training and rendering NeRF models, and select nerfacto and splafacto that are the basic models supported in NerfStudio. The nerfacto model, referred to as ``NeRF'' in the remainder of this section, is a deep neural network model balanced between speed and quality. Originating from 3D-GS~\cite{kerbl20233d}, the splafacto model, referred to as ``GS'', is an implementation of the 3D-GS with GPU-backed rasterization for rapid rendering. Both are within 6 GB of memory, suitable for edge devices like NVIDIA Jetson.

Figure~\ref{fig:robotdataset} shows the manually collected dataset used in the benchmarking. The scene is captured with 40 6-megapixel RGB images, with the focus around the robot arm as the target object. We use COLMAP~\cite{schoenberger2016sfm} to calculate their perspective coordination in the space by feeding the images that share regions in the image.

\subsection{Computing Benchmarking}
The training runs for 30,000 iterations while measuring GPU utilization and elapsed training time. In the end, we evaluate the performance of the trained models using the `ns-eval' command in NerfStudio. As shown in Figure~\ref{fig:roboteval}, the evaluation focuses on the quality of view synthesis against the ground-truth image used in the training.

\subsubsection{Evaluation Metrics}
Table~\ref{tab:benchmark} reports six key metrics across all platform-method combinations. We briefly define each:

\begin{itemize}
\item \textbf{PSNR $\uparrow$} (Peak Signal-to-Noise Ratio, in dB): Measures reconstruction fidelity by comparing pixel-level similarity between rendered and ground-truth images. Higher values indicate better visual quality. PSNR is standard in 3D reconstruction but saturates at high quality thresholds.
\item \textbf{SSIM $\uparrow$} (Structural Similarity Index): Captures perceived structural similarity between rendered and ground-truth images, addressing PSNR's perceptual gaps. Values range from 0 to 1; higher is better.
\item \textbf{LPIPS $\downarrow$} (Learned Perceptual Image Patch Similarity): A deep-learning-based metric that aligns better with human perceptual judgment than PSNR or SSIM. Lower values indicate better perceived visual quality.
\item \textbf{FPS $\uparrow$} (Frames Per Second): Rendering throughput during inference; critical for real-time perception loops. Values below 10 FPS represent unacceptable latency for interactive robotic tasks.
\item \textbf{Time (minute)}: Elapsed wall-clock training time in minutes; reflects absolute time to obtain a trained model.
\item \textbf{GPU-minute}: Effective GPU compute cost, calculated as elapsed time $\times$ averaged GPU utilization, normalizing for hardware idling. Allows fair comparison across devices with different utilization profiles.
\end{itemize}

\begin{figure*}[tb]
\centering
  \begin{subfigure}[b]{0.16\textwidth}
    \centering
    \includegraphics[width=\textwidth]{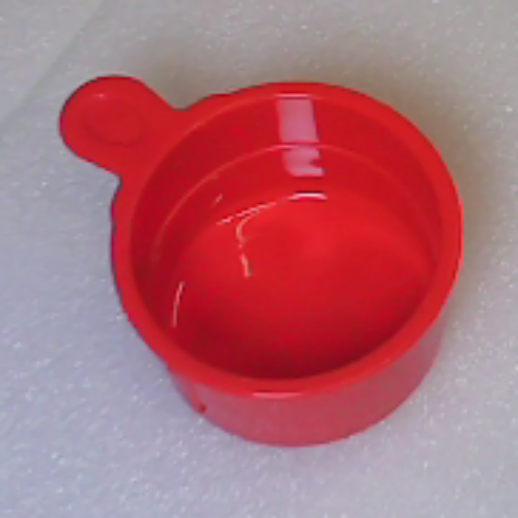}
  \end{subfigure}
  \begin{subfigure}[b]{0.16\textwidth}
    \centering
    \includegraphics[width=\textwidth]{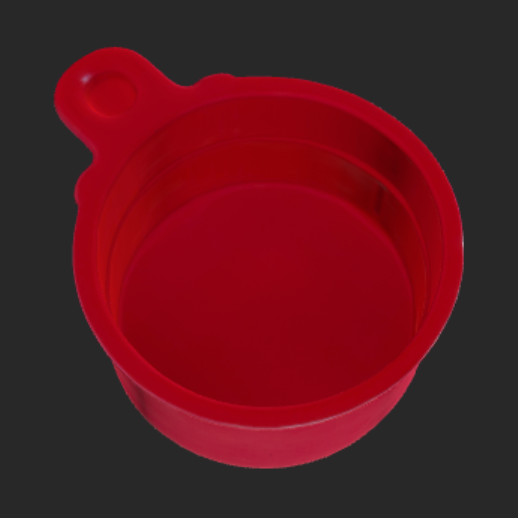}
  \end{subfigure}
  \hfill
  \begin{subfigure}[b]{0.16\textwidth}
    \centering
    \includegraphics[width=\textwidth]{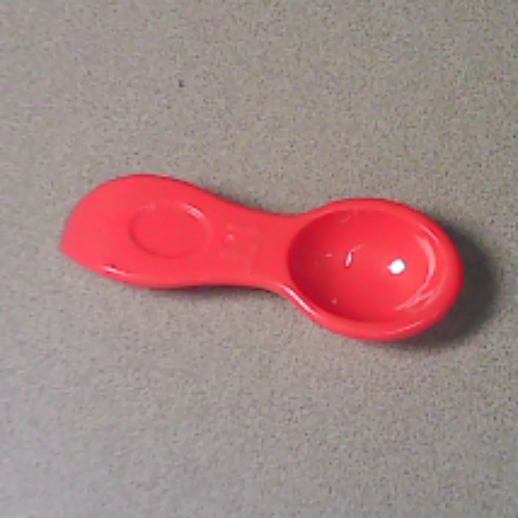}
  \end{subfigure}
  \begin{subfigure}[b]{0.16\textwidth}
    \centering
    \includegraphics[width=\textwidth]{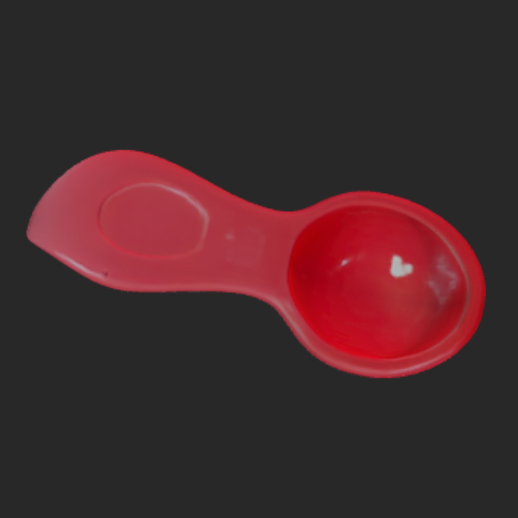}
  \end{subfigure}
  \hfill
  \begin{subfigure}[b]{0.16\textwidth}
    \centering
    \includegraphics[width=\textwidth]{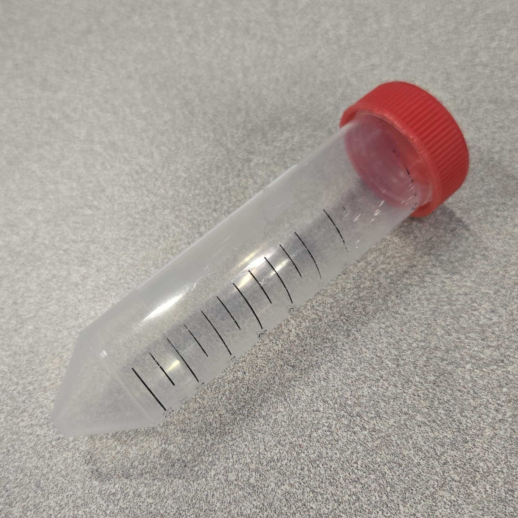}
  \end{subfigure}
  \begin{subfigure}[b]{0.16\textwidth}
    \centering
    \includegraphics[width=\textwidth]{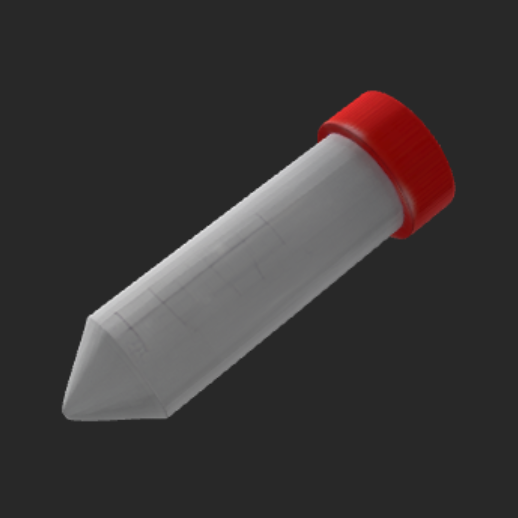}
  \end{subfigure}
  \vfill
  \bigskip
  \begin{subfigure}[b]{0.16\textwidth}
    \centering
    \includegraphics[width=\textwidth]{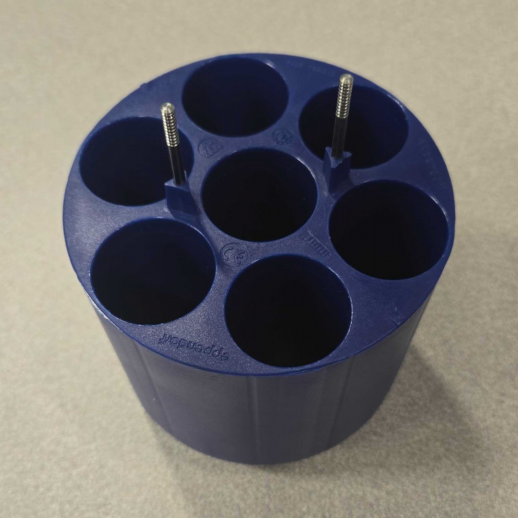}
  \end{subfigure}
  \begin{subfigure}[b]{0.16\textwidth}
    \centering
    \includegraphics[width=\textwidth]{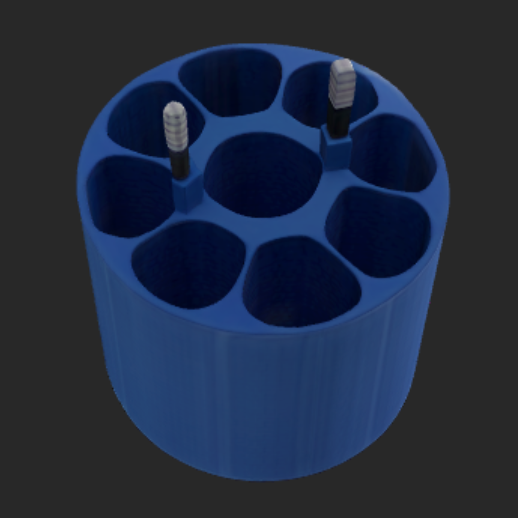}
  \end{subfigure}
  \hfill
  \begin{subfigure}[b]{0.16\textwidth}
    \centering
    \includegraphics[width=\textwidth]{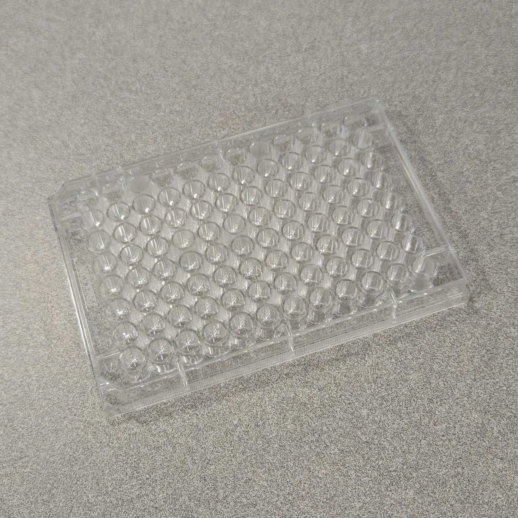}
  \end{subfigure}
  \begin{subfigure}[b]{0.16\textwidth}
    \centering
    \includegraphics[width=\textwidth]{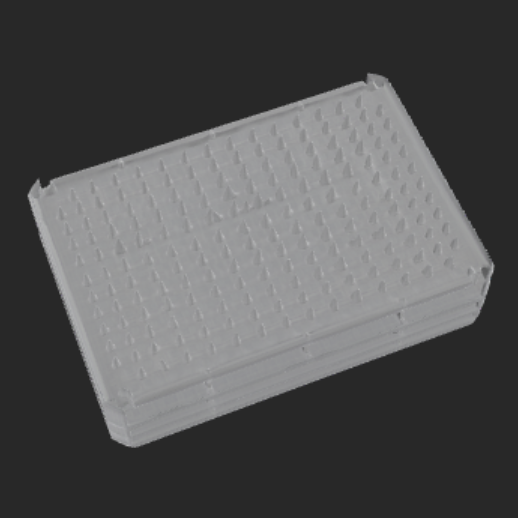}
  \end{subfigure}
  \hfill
  \begin{subfigure}[b]{0.16\textwidth}
    \centering
    \includegraphics[width=\textwidth]{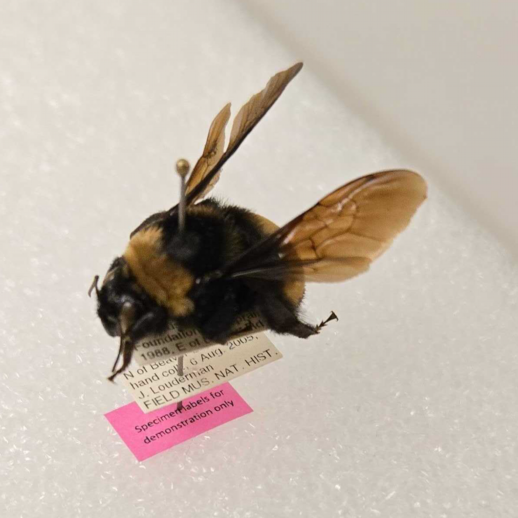}
  \end{subfigure}
  \begin{subfigure}[b]{0.16\textwidth}
    \centering
    \includegraphics[width=\textwidth]{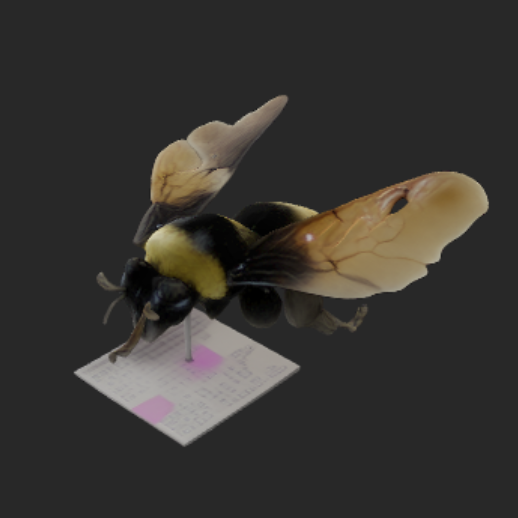}
  \end{subfigure}
  % \hfill
  % \begin{subfigure}[b]{0.33\textwidth}
  %   \centering
  %   \includegraphics[width=\textwidth]{figures/eval_gs.png}
  %   \label{fig:device_sophia}
  % \end{subfigure}
  \caption{Reconstructed 3D models from SAM3D. Each pair consists of an input image on the left and a generated 3D model on the right. From top-left to bottom-right: toy measuring cup, toy spoon, plastic tube with cap, centrifuge bucket, pipette well plate, and taxidermy bee.}
  \label{fig:sam_objects}
\end{figure*}

\subsubsection{Key Findings}
Table~\ref{tab:benchmark} shows the results of NeRF and GS benchmarking on the computing devices. In all devices, the GS method yields higher PSNR and faster rendering than NeRF. However, it tends to train longer and consume more GPU resources. For instance, the GPU-minute --- the wall-clock minutes for GPU computation calculated from \textit{elapsed time} $\times$ \textit{averaged GPU utilization} --- of GS on Alienware takes roughly 1.4x more than NeRF on the same device.

\textit{Quality-vs.-Cost Trade-off:} GS achieves 5.95 PSNR points higher than NeRF on the Sophia HPC node (24.32 vs 18.96 dB), at the cost of 70\% more GPU-minutes (13.8 vs 8.1). This suggests that the higher fidelity comes at a substantial compute premium. On the Jetson Orin—the platform most likely to be mounted onboard a laboratory robot---GS still outperforms NeRF in quality but requires 25\% more GPU-time (65.3 vs 52.3 GPU-minutes), a trade-off that becomes critical when onboard power budgets are constrained to \textasciitilde50 W.

\textit{Rendering Latency:} GS achieves higher FPS over NeRF because GS uses hardware-accelerated and optimized rasterization for reconstruction. We suspect that Alienware and Orin devices benefit more from this optimization due to their GPU architectures than Sophia's A100. For interactive robotic manipulation, this result indicates that real-time perception loops must rely on Gaussian splats for reconstruction and rendering, with periodic refinement through heavier neural methods.

\textit{Platform Scaling:} Across the devices, the Sophia node shows the fastest training as expected, followed by the Alienware desktop computer and the Jetson Orin. The elapsed training time on the Orin has a huge gap from the other two, indicating that Orin's insufficient computing power prevents it from supporting the task effectively. Specifically, training a single GS model takes 16.5 minutes on Sophia, 21.8 minutes on Alienware, and 90.0 minutes on the Jetson Orin—a 5.5x slowdown relative to the server-class platform. This scaling behavior is critical for pipeline design: onboard optimization is prohibitively slow, necessitating edge or cloud-based reconstruction with periodic synchronization.

Note that we utilized only 1 out of 8 available A100 GPUs in Sophia. Using PyTorch's Distribution Data Parallel (DDP) and/or model-split methods~\cite{li2024nerf} can further accelerate the training on multi-GPU devices. Such multi-GPU strategies could significantly reduce Sophia's training time, reinforcing the gap between cloud and onboard reconstruction capabilities.

\section{Preliminary result of SAM3D reconstruction}
To understand the performance of LRMs for single-image inference, we test SAM3D with single images taken from a camera. We downsize the input image into 512 x 512 and feed it once into the SAM3D inference running on a desktop-grade computer with an NVIDIA RTX 5000 GPU. The inference takes roughly 9-12 seconds to generate a 3D model; generating Gaussian splats from SAM3D takes 8-10 seconds, while mesh decoding finishes within a second.

\subsection{Object Reconstruction Examples}

Figure~\ref{fig:sam_objects} shows example 3D models generated from SAM3D. These models follow the shape of the object in the input image, and 3D shapes in the occluded area also resemble the example objects, which appear symmetric. However, we see challenging examples where the object geometry is not precisely reconstructed in the 3D model (see the bottom row in Fig.~\ref{fig:sam_objects}). The centrifuge bucket has 7 openings for accepting plastic tubes, but the associated 3D model has more openings, which can cause misalignment of the tube into the opening when robots refer to the 3D model for the manipulation. The same problem occurs in the well plate, with the wrong number of openings. Lastly, the 3D bee model does not seem to capture the shape and volume that are unique to the bee.

% note for evaluation
% Ref: Tanks and Temples: Benchmarking Large-Scale Scene Reconstruction
% Alignment. For benchmarking, the point clouds produced by the
% evaluated pipelines must be aligned to the ground-truth models.
% Most pipelines expose the reconstructed camera poses, and for these
% we perform the alignment automatically. The reconstructed camera poses are registered to estimated ground-truth camera poses,
% yielding scale and pose estimates for the reconstructed point cloud.
% We estimate the ground-truth camera poses using the ground-truth
% point cloud [Mastin et al. 2009]. Alternatively, for pipelines that do
% not expose the camera poses (e.g., Pix4D), we manually align the
% reconstructed point cloud to the ground truth.

\subsection{Assessing Quality of SAM3D Reconstruction}
\begin{figure}[tb]
\centering
  \begin{subfigure}[b]{\columnwidth}
  \includegraphics[width=\columnwidth]{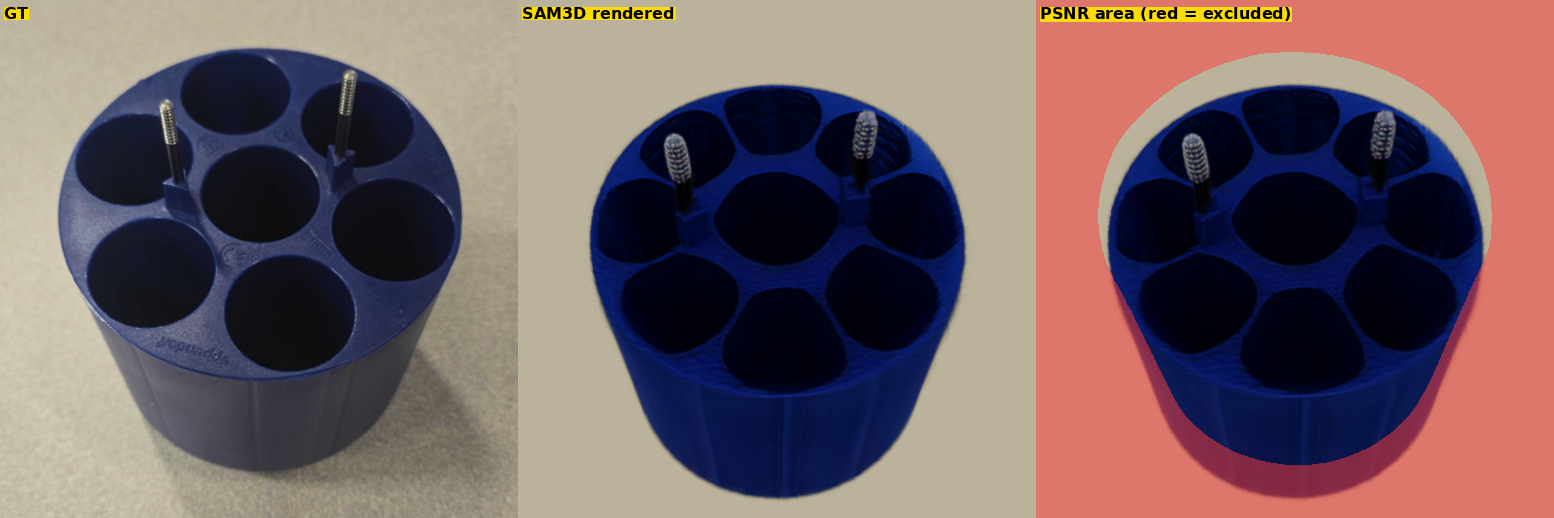}
  \end{subfigure}
  \vfill
  \bigskip
  \begin{subfigure}[b]{\columnwidth}
  \includegraphics[width=\columnwidth]{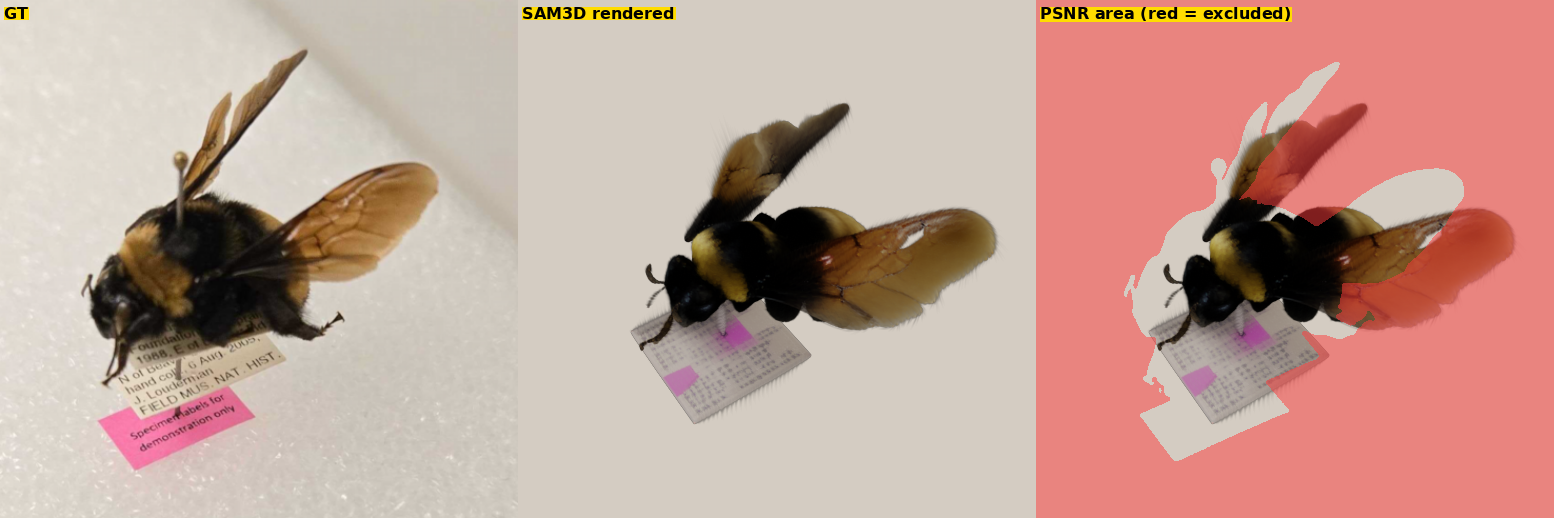}
  \end{subfigure}
  \caption{(left) A ground-truth image, (middle) SAM3D reconstruction from the image, and (right) masked image showing the area for PSNR calculation. The red area is excluded for the calculation. The scale and pose mismatch between the image and the reconstructed model make PSNR calculation challenging.}
  \label{fig:sam-visual-quality}
\end{figure}

Although the reconstructed 3D models appear reasonably shaped, we find two problems in comparing the visual quality of the reconstructed object against the ground-truth image. First, different from NeRF and GS, the ground-truth image misses its pose information to match the perspective of the reconstructed model to the physical object. Heuristically estimating the exact pose of the ground truth image is not feasible. As a consequence, simply comparing PSNR to gauge the visual quality of reconstructed 3D model faces this matching problem (see Fig.~\ref{fig:sam-visual-quality}). Secondly, SAM3D constructs a 3D object model without rendering the scene, and hence the ground-truth image needs to be background removed. Background removal is currently done either by masking the object of interest in SAM3 or by using automated background removal tools such as RemBG. However, if the target object is in a complex scene, background removal may leave some background in the image and may cause false positives in the region, negatively impacting PSNR calculation.

To evaluate the geometric and volumetric quality of reconstructed objects, many studies use the Chamfer distance and the F-1 score. SAM3D respectively reports their F-1 score and Chamfer distance as 0.2344 and 0.04, on the SA-3DAO dataset (see the detailed comparison in ~\cite{chen2026sam}). Despite of these scores, our preliminary assessment in this zero-shot reconstruction raises concerns in the details of reconstructed object models in the context of automated robot manipulation of these models due to the detail mismatch. This necessitates continuous model finetuning with laboratory objects as well as any unseen, unmodeled object that would be discovered during laboratory experiement.

\section{Remark and future work}
3D reconstruction and rendering are essential components for autonomous robots to understand the surrounding environment and plan actions. The NeRF and GS methods provide high-quality view synthesis but require multi-perspective images and a sufficient amount of training time. On the other hand, transformer-based models quickly produce Gaussian splats to generate a 3D model, but the visual and detailed geometric quality for autonomous robot tasks is questionable at the early stage of our assessment. We see that the latency-fidelity balance exists, and that can be managed when we scale computation at every level in the infrastructure, spanning from mobile robots, scientific instruments, edge computers, to HPC systems.

\subsection{Benchmark Scope and Limitations}
This work focuses narrowly on the computational and reconstruction-quality aspects of neural 3D reconstruction methods across compute platforms. We do not propose new reconstruction techniques, nor do we develop visual analytics capabilities for system-level decision support.  Instead, we provide empirical benchmarking data to inform engineering choices in pipeline design. Robotic manipulation tasks, particularly those in laboratory settings, impose strict latency requirements on perception loops. Open datasets~\cite{Ego4D2022CVPR, khazatsky2024droid} for visual-language-action models record visual data at 15 to 20 Hz, implying that the perception loop should operate ideally at a similar frame rate to sustain real-time interaction.

\subsection{Direct Future Work: Reconstruction Methods}
Future work includes performance benchmarking of LRM models for lab-related objects. Obtaining precisely 3D-scanned objects can help with benchmarking, in addition to using open datasets. On the GS model training, we will explore ways to speed up the training by importing Gaussian splats generated from LRM models. This approach can not only loosen the need for multi-perspective images to train the GS model but also enable adaptive strategies to generate high-fidelity splats on focused parts of the object.

\subsection{Future Work: Visual Analytics and Science-Aware Autonomy}
The benchmarking results presented here establish a foundation for designing reconstruction pipelines; however, integrating these pipelines into a visual analytics system for science-aware autonomous manipulation is left as future work. Such a system would need to: (1) visualize reconstruction outputs to enable scientist monitoring and intervention; (2) track reconstructed models with their scientific context (e.g., knowing what chemicals are synthesized in the flask); (3) provide interactive decision support for autonomous platform to select between latency-optimized and fidelity-optimized reconstruction strategies based on task requirements; and (4) integrate these systems into the full autonomous laboratory workflow. These capabilities require domain-specific design, user study validation, and integration with laboratory automation platforms—efforts beyond the scope of this compute-platform benchmark.

\bibliographystyle{abbrv-doi}

\bibliography{template}
\end{document}